\documentclass[lettersize,journal]{IEEEtran}
\usepackage{amsmath,amsfonts}
\usepackage{algorithmic}
\usepackage{algorithm}
\usepackage{array}
\usepackage[caption=false,font=normalsize,labelfont=sf,textfont=sf]{subfig}
\usepackage{textcomp}
\usepackage{stfloats}
\usepackage{url}
\usepackage{verbatim}
\usepackage{graphicx}
\usepackage{cite}
\usepackage{multirow}
\usepackage[colorlinks=true, allcolors=blue]{hyperref}

\usepackage{threeparttable}
\usepackage{soul}
\sethlcolor{yellow}

\IEEEpubid{\begin{minipage}{\textwidth}\ \centering
		Copyright \copyright 2025 IEEE. Personal use of this material is permitted. \\
		However, permission to use this material for any other purposes must be obtained 
		from the IEEE by sending an email to pubs-permissions@ieee.org.
\end{minipage}}

\begin{document}

\title{Meta knowledge assisted Evolutionary Neural Architecture Search}

\author{Anonymous Author(s)}

\author{Yangyang~Li,~\IEEEmembership{Senior~Member,~IEEE,} 
	Guanlong~Liu, 
	Ronghua~Shang,~\IEEEmembership{Senior~Member,~IEEE,} 
	and~Licheng~Jiao,~\IEEEmembership{Fellow,~IEEE}%
    \thanks{Yangyang Li, Guanlong Liu, Ronghua Shang, and Licheng Jiao are with the Key Laboratory of Intelligent Perception and Image Understanding of Ministry of Education, Joint International Research Laboratory of Intelligent Perception and Computation, International Research Center for Intelligent Perception and Computation, Collaborative Innovation Center of Quantum Information of Shaanxi Province, School of Artificial Intelligence, Xidian University, Xi'an 710071, China (e-mail: 
		\href{mailto:yyli@xidian.edu.cn}{\textcolor{black}{yyli@xidian.edu.cn}}; 
		\href{mailto:gloliu@stu.xidian.edu.cn}{\textcolor{black}{gloliu@stu.xidian.edu.cn}}; 
		\href{mailto:rhshang@mail.xidian.edu.cn}{\textcolor{black}{rhshang@mail.xidian.edu.cn}}; 
		\href{mailto:lchjiao@mail.xidian.edu.cn}{\textcolor{black}{lchjiao@mail.xidian.edu.cn}}).}%
	\thanks{This work was supported by the National Natural Science Foundation of China under Grant 62476209, the Key Research and Development Program of Shaanxi under Grant 2024CY2-GJHX-18, the Research Project of SongShan Laboratory under Grant YYJC052022004, and the Natural Science Basic Research Program of Shaanxi under Grant No. 2022JC-45. (Corresponding author: Guanlong Liu.)}
	
}

%

\maketitle

\begin{abstract}

Evolutionary computation (EC)-based neural architecture search (NAS) has achieved remarkable performance in the automatic design of neural architectures. However, the high computational cost associated with evaluating searched architectures poses a challenge for these methods, and a fixed form of learning rate (LR) schedule means greater information loss on diverse searched architectures. This paper introduces an efficient EC-based NAS method to solve these problems via an innovative meta-learning framework. Specifically, a meta-learning-rate (Meta-LR) scheme is used through pretraining to obtain a suitable LR schedule, which guides the training process with lower information loss when evaluating each individual. An adaptive surrogate model is designed through an adaptive threshold to select the potential architectures in a few epochs and then evaluate the potential architectures with complete epochs. Additionally, a periodic mutation operator is proposed to increase the diversity of the population, which enhances the generalizability and robustness. Experiments on CIFAR-10, CIFAR-100, and ImageNet1K datasets demonstrate that the proposed method achieves high performance comparable to that of many state-of-the-art peer methods, with lower computational cost and greater robustness.

\end{abstract}


\section{Introduction}\label{section:1}
\label{sec: 1}
\IEEEPARstart{T}{he} accelerated development of deep learning has achieved remarkable success in many challenging real-world tasks, such as image classification \cite{niepert2016learning}, object tracking \cite{zhao2019object,jiao2021deep} and natural language processing \cite{otter2020survey}. Neural networks, such as ResNet \cite{he2016deep} and MobileNetV2 \cite{sandler2018mobilenetv2}, are always derived from considerable manual experience. Nevertheless, some neural networks are only applicable to a single dataset, and the manually crafted architectures often present challenges for common users, as they require the expertise of human specialists in both the deep neural networks (DNNs) and the relevant problem domain to design effectively. Recently, automated machine learning (AutoML) has gained increasing attention from both industry and academia \cite{kim2022neural,xue2022partial,lu2023neural}. Neural architecture search (NAS), a cutting-edge technology in AutoML, aims to automate the design of DNNs with better performance, especially under complex constraints in real-world tasks.

In general, there are numerous methods that can be used for NAS. For example, reinforcement learning (RL)-based methods \cite{zoph2018learning,vahdat2020unas,lyu2021multiobjective}, gradient descent (GD)-based methods \cite{liu2018darts,chen2024automated,tan2024universal}, and evolutionary computation (EC)-based methods \cite{wen2021two,guan2022large,zhang2023fast} are popular in architecture search. RL-based algorithms construct the model via a reward function, which is designed on the basis of an evaluation of the model's performance. While this type of method can achieve high accuracy, it is often extremely computationally expensive. For example, on the widely used small natural image dataset CIFAR-10, NAS-RL \cite{zoph2018learning} requires simultaneously running for thousands of GPU days, which poses an insurmountable challenge for the majority of researchers and devices. GD-based algorithms consider architecture search to be a continuous optimization problem \cite{feng2024lrnas}, which can search more efficiently because of the continuously differentiable operation space. However, algorithms such as DARTS \cite{liu2018darts} always exhibit unstable performance, and their search space is restricted by the predefined operation space.

EC-based algorithms have received extensive attention because of their ability to solve nondifferentiable problems as well as their global search ability to handle complex constraints and optimization problems. Real et al. \cite{real2017large} introduced a large-scale evolution framework to explore novel architectures spanning 2750 GPU days. In \cite{sun2020automatically}, Sun et al. proposed a method named CNN-GA for the automatic design of convolutional neural network (CNN) architectures using genetic algorithms. Although it took up to 35 GPU days to find the best architecture, it helped people without domain knowledge of the CNN obtain promising CNN architectures for a given image. Sun et al. \cite{sun2020completely} proposed AE-CNN, a CNN architecture that evolved by incorporating two valid blocks. It took approximately 27 GPU days to obtain an optimized CNN structure on CIFAR-10 \cite{krizhevsky2009learning}. NSGANet, which explores the space of feasible architectures through a three-step process and can be transferred to multiple scenarios, was proposed by Lu et al. \cite{lu2019nsga}. As a result, it discovered an architecture that achieved an accuracy rate of 97.25\% using 4 GPU days. Zhou et al. \cite{zhou2023towards} proposed a knowledge-sharing mechanism in multitask scenarios and achieved \IEEEpubidadjcol competitive performance with a 2$\times$ lower search cost. Zhao et al. \cite{zhao2024knowledge} presented a knowledge-guided adaptive multiobjective evolutionary strategy for remote sensing classification that effectively combines the strengths of CNN and Transformer via the knowledge learned from natural images. \cite{zhang2024boosting} proposed a concept of supernet shifting to improve order-preserving ability, but it focuses primarily on fine-tuning the supernet.
To address the major computational cost limitation, significant efforts have been made to increase computational efficiency using multiple techniques, including weight inheritance \cite{zhang2020efficient,huang2023split}, weight sharing \cite{zhang2023evolutionary,hu2022pwsnas}, surrogate model \cite{lu2020nsganetv2,wei2022npenas,zhou2023surrogate}, and population reduction \cite{wen2021two}.

An appropriate learning rate (LR) schedule can significantly improve a model's generalizability and DNN training results. Wu et al. \cite{wu2018sgd} also theoretically highlighted the unique role of the LR schedule in global minimum selection. Meta-learning enables models to automatically acquire appropriate learning rates by optimizing the adaptation process across multiple tasks. Instead of manually tuning fixed learning rates, the meta-learning framework meta-SGD \cite{li2017meta} treats the learning rate itself as a learnable parameter within a bilevel optimization process. The meta-learner optimizes the learning rate across tasks by minimizing a meta-objective. The model then performs task-specific updates using gradient descent with a learning rate guided by the meta-learner. While the meta-learning mechanism can make the model more suitable for the data and significantly improve the model's generalizability and robustness, traditional meta-learning strategies are usually computationally expensive and focus on optimizing specific neural architectures, which makes applying meta-learning to NAS difficult.

To solve the abovementioned problems, we design a novel and efficient EC-based NAS method through a meta-learning framework, termed MetaNAS. The proposed method achieves a good balance between computational cost and model performance by designing an efficient learning rate (LR) schedule in advance during pretraining, utilizing a meta-learning rate (Meta-LR) scheme. Each individual is then evaluated using the tailored LR schedule. Unlike traditional meta-learning algorithms, which always take the gradients as input, the Meta-LR scheme learns only the LR schedule, which avoids the shortcomings of traditional meta-learning algorithms, which are computationally intensive and difficult to generalize to different architectures. The Meta-LR scheme can be easily used in existing EC-based NAS methods to increase their robustness and reduce complexity. We introduce an adaptive surrogate model with an adaptive threshold, which can predict potentially superior architectures in a few epochs and evaluate the selected architectures with complete epochs. In addition, we design a period mutation operator in the evolutionary algorithm to increase population diversity and improve the generalizability, making it more in line with the current trend of accelerating NAS methods.

In conclusion, the key contributions of this paper are summarized as follows:
\begin{enumerate}[]
	\item 	We present a novel meta-learning framework to improve the generalizability and robustness of NAS across various architectures by combining meta-learning with performance evaluation in the search process. The LR schedule can be dynamically adjusted on the basis of training loss with the Meta-LR scheme during pretraining so that the acquired prior knowledge can be effectively applied to evaluate the potential architectures for complete epochs without extra computational cost.
	\item 	To improve the search efficiency, an adaptive surrogate model is incorporated into the evaluation by applying an adaptive threshold. The threshold flexibly determines whether to evaluate the architecture with complete epochs. In this way, potential architectures encoded by individuals can be found more efficiently, which greatly reduces the computational cost of the evolution algorithm. 
	\item	We introduce a customized period mutation genetic operator to address the acceleration of NAS by periodically adjusting the mutation probability. The presented genetic operator can help find more potential and diverse architectures and further underscores the enhanced generalization ability of the searched architecture. 
	\item 	The effectiveness of the proposed method is validated on three popular image datasets: CIFAR-10, CIFAR-100 and ImageNet1K. Extensive experiments and ablation studies prove that the proposed MetaNAS performs favorably compared with state-of-the-art methods.
\end{enumerate}

The rest of this paper is structured as follows. Section II provides a brief overview of related work. Section III elaborates on the proposed MetaNAS and its algorithm. The comparison experiments and discussion of the ablation results are presented in Section IV. Finally, Section V presents the conclusion of this paper.

\section{Related Works}\label{section:2}
\label{sec: 2}
In this section, preliminary work on EC-based NAS is first briefly introduced. We then review current research on the reduction in NAS computational cost. The motivation for combining meta-learning and NAS is presented at the end.

Research on NAS can generally be broadly divided into the following three aspects: search algorithms, search spaces, and evaluation strategies. Typical search algorithms include RL-based, GD-based and EC-based methods. The RL-based NAS method was first proposed, in which the representation and optimization of the agent is the key to applying reinforcement learning to search for the neural architecture. Because RL-based methods always consume many computing resources, it is difficult to use these methods in real-world applications. The GD-based methods \cite{liu2018darts,zhang2020you,peng2024recnas} can greatly reduce the computational cost and improve the search speed by making the search space continuous and using gradients to optimize the model parameters and network structure weights alternately. However, these methods do not guarantee convergence in their optimization algorithms, which means that parameter tuning is a serious challenge. In fact, GD-based algorithms often find ill-conditioned architectures because of the incorrect relationship for adapting to alternate gradient descent optimization \cite{liu2023survey}. Additionally, they require a high level of expertise to construct a superb network in advance.
EC-based \cite{zhang2020efficient,zhang2023evolutionary,yang2022accelerating} methods typically use discrete search spaces, which provide better generalizability for the evaluation strategy. As evolutionary algorithms are insensitive to local minima, EC has been widely applied to solve complex nonconvex optimization problems \cite{sun2018igd}. In real-world applications, there is a high requirement for network scalability and inference latency, so model complexity is often considered in NAS tasks \cite{zhou2021survey, singh2024kl}. 

\subsection{Evolutionary Computation-based Neural Architecture Search}\label{section:2.1}
Compared with GD-based methods, EC-based NAS methods tend to save more computational resources than the methods based on RL do and do not require supernet training. In addition, EC can solve constrained optimization problems \cite{li2017overview}. EC-based methods can search for promising models on the basis of constraints. Mathematically, a bi-objective optimization problem can be used to express NAS as follows:
\begin{equation}	
	\min_{x}F(x) =
	\begin{cases}
		f_1(x)=1-f_{accuracy}(x,D_{val}) \\f_2(x)=f_{complexity}(x)
	\end{cases}
\end{equation}
where $x$ represents the searched architecture, $f_1(x)$ represents the error rate of $x$ on the validation dataset $D_{val}$, and $f_2(x)$ denotes the complexity of $x$. The complexity can be quantified on the basis of various factors, such as Floating Point Operations (FLOPs), the model parameter size and inference latency, to identify neural architectures with varying complexities to suit diverse deployment scenarios. 

Researchers have suggested the Pareto front framework to evaluate the performance of structures effectively in a comprehensive manner. To address the multiobjective architecture search problem, various algorithms have been proposed. Some studies add inference delay as a constraint to search for more robust structures under resource-constrained conditions and obtain higher classification accuracy. Lu et al. \cite{lu2019nsga} considered FLOPs as an additional objective. FLOPs can be used as a constraint of the algorithm to realize reduced resource consumption and improved accuracy. 

\subsection{Cost--Performance Balance}\label{section:2.2} 
To address the high computational cost, existing EC-based NAS methods can be broadly categorized into two groups:

1)	Methods with a lower evaluation cost for each architecture.

\textbf{Weight Inheritance.} Weight inheritance is a strategy for replicating and inheriting the weights of an existing parent architecture in a child architecture, thus greatly utilizing known information and improving search efficiency. A sampled training strategy implemented in this approach trains all individuals at the beginning of each iteration \cite{zhang2020efficient}. This ensures that each parent is equipped with the necessary knowledge and experience before offspring individuals are generated. During each iteration, the offspring individuals inherit weights from the parent individuals. This way, the offspring do not require any further training on the basis of their weights, which decreases computational cost and the amount of resources that would otherwise be needed for additional training. Huang et al. \cite{huang2023split} developed a unique method for accelerating architecture evaluation called elite weight inheritance. They constructed an online external weight pool to save time. For each layer, the core idea is to select the best operations. 

\textbf{Weight Sharing.} Weight sharing is the process of obtaining the weights of the architecture directly from a group of weights stored in the pretrained supernet model. Oneshot NAS has emerged as a promising approach to reduce computational cost. It can obtain a variety of different networks for different constraints. Therefore, the candidate architectures can be directly evaluated by leveraging the weights inherited from the pretrained supernet, eliminating the need for further training. The proposed partial weight sharing method discussed in \cite{zhang2023evolutionary} effectively limits the degree of weight sharing through the help of a customized crossover operator, thereby reducing the adverse impacts of weight coupling in Oneshot NAS. Hence, the algorithm can search for several diverse subnets. By automatically shrinking the search space, Hu et al. \cite{hu2022pwsnas} proposed a framework that results in a smaller promising search space by progressively simplifying the original search space.

2)	Methods yielding fewer architectures to be evaluated.

\textbf{Surrogate Model.} In general, the primary concept behind employing a surrogate model is to accurately forecast the quality of architectures represented by promising individuals, which allows for a reduction in the number of individuals who require evaluation. Lu et al. \cite{lu2020nsganetv2} introduced a bilevel surrogate model comprising two different surrogates. One upper-level surrogate enhances sample efficiency. The lower-level surrogate facilitated by a pretrained supernet is used to improve weight learning efficiency. In \cite{wei2022npenas}, two different graph-based predictors were developed to guide EA in enhancing exploration ability. Zhou et al. \cite{zhou2023surrogate} used a random forest to assist in decomposing the high-dimensional search space.

\textbf{Population Reduction.} A recursive decrease in population size can achieve EC-based NAS acceleration. In \cite{wen2021two}, a two-stage EA with different population sizes in each stage was proposed, where a coarse search was performed to identify suitable candidate neural architectures and a local fine-grained search was performed for refinement.

\subsection{Meta-Learning in Architecture Search}\label{section:2.3}
Meta-learning has been used in various ways in deep learning and is commonly understood as $\mathit {learning\ to\ learn}$. There are three basic problems in meta-learning: how meta-knowledge characterizes knowledge, how the meta-learner selects an optimizer for optimization and learning, and the meta-objective answers the question of why learn this way in meta-learning. 
In fact, meta-learning expects the inner learning algorithm itself to tune parameters with an outer meta-learner so that it can quickly transfer to new tasks on the basis of existing knowledge rather than training from scratch. In other words, the meta-learner has generalizability and robustness. A series of meta-learning algorithms was developed by J. Schmidhuber, and the concept of self-directed learning was used in these methods \cite{schmidhuber1987evolutionary}. S. Thrun first created and explained the notion of $\mathit {learning\ to\ learn}$ in \cite{thrun1998learning} and represented it as a way to implement meta-learning. In recent years, meta-learning has focused on learning common knowledge shared from invisible tasks, with the expectation that the learned knowledge can be transferred to abstruse tasks \cite{shu2023cmw}. In addition, some works have employed meta-learning for a wide range of optimization problems, such as evolution strategies \cite{houthooft2018evolved}, combinatorial functions \cite{rosenfeld2018learning}, and few-shot learning \cite{chen2017learning}.

In NAS tasks, a fixed form of the LR schedule results in greater information loss on diverse searched architectures. There are many possible combinations of operations in the search space, and diverse architectures have difficulty adapting to complex optimization problems via hand-designed LR schedules. Although these methods may perform better in certain cases, the optimizers learned by these methods often struggle to work in different situations. As stochastic gradient descent (SGD) \cite{robbins1951stochastic,polyak1964some} was introduced, massive adaptive LR methods based on gradient descent algorithms, such as Adagrad \cite{duchi2011adaptive}, RMSprop \cite{tieleman2012lecture}, and Adam \cite{kingma2014adam}, have been developed for the deep learning domain. 
Adaptive LR methods allow for quicker convergence, especially in scenarios where the loss surface is complex and the gradient varies significantly across dimensions. In general, these optimizers prevent unnecessary oscillations or slow progress in flat zones by adjusting the learning rates over time. However, methods such as Adam, which requires approximately twice the memory used by SGD, may result in higher computational cost. 

Similarly, the LR schedule obtained by the Meta-LR scheme can adapt well to new architectures.
Considering that optimizers \cite{chen2017learning} that use gradients as input and output cannot generalize over different network structures and have difficulty coping with large-scale optimization problems \cite{shu2022mlr}, the proposed method performs a meta-learning scheme based on the input training loss when evaluating the architectures during the search process and dynamically optimizes the LR.

\section{Methodology}\label{section:3}
\label{sec: 3}
\begin{figure*}[htbp]
	\centering
	\includegraphics[width=\linewidth]{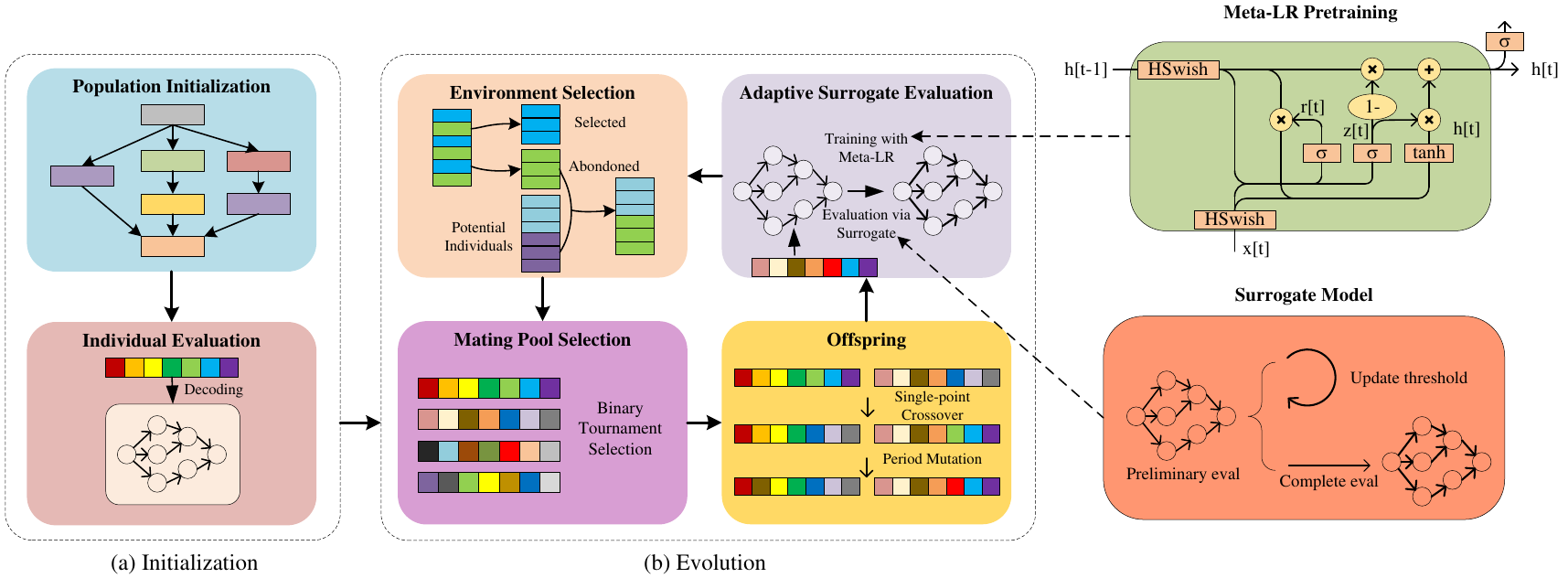}
	\caption{ Overall diagram of the proposed MetaNAS, which combines six steps. We first obtain the individuals by initializing the population. The proposed MetaNAS framework is subsequently used to find the individuals with the best fitness to decode the best architecture.}
	\label{fig:1}
\end{figure*}

Traditional meta-learning algorithms use two-layer optimization, where numerous internal steps need to be stored in memory, which is expensive in terms of time and memory. Therefore, this paper proposes a novel MetaNAS framework to solve the problem. An appropriate learning rate is obtained by self-learning in pretraining as the meta-knowledge, and the evolutionary algorithm subsequently utilizes the adaptive surrogate model as the meta-learner. Finally, the fitness score of each architecture is used as the meta-objective to find the potential architectures. The overall pipeline is shown in Fig. \ref{fig:1}, and below, we provide a thorough description of the proposed MetaNAS.

\subsection{Overall Framework}\label{section:3.1}
As observed in Fig. \ref{fig:1}, MetaNAS comprises six steps. The detailed steps are shown in Algorithm \ref{alg:1}.
\begin{algorithm}[htb]
	\caption{Framework of the proposed MetaNAS}
	\label{alg:1}
	\renewcommand{\algorithmicrequire}{ \textbf{Input:}}     
	\renewcommand{\algorithmicensure}{ \textbf{Output:}}    
	\begin{algorithmic}[1] 
		\REQUIRE Dataset $D$, Maximum number of evolutionary generations $G$, Population Size $P_s$, Training Epochs $M$, Threshold of top1 error $H_t$.
		\ENSURE Final Population $P$.
		
		\STATE /* Stage 1: Initialization */
		\STATE $P$, $H$ $\gets$ Initialization($P_s$);
		\STATE Evaluate individuals in $P$ by training architectures encoded by the individuals for $M$ epochs and compute the fitness;
		
		\STATE /* Stage 2: Evolution */	
		\FOR {$j = 1,…, {G}$}
		\WHILE {termination criterion is not satisfied}	
		\STATE $P{'}$ $\gets$ Select $P_s$ parents according to the non-dominated front number and crowding distance of individuals in $P$;
		\STATE $Q$ $\gets$ Genetic Operator($P{'}$);
		\STATE $H$ $\gets$ Evaluate offspring $Q$ by training architectures with the LR schedule obtained by Meta-LR scheme and predict top1 error via surrogate model; /* Algorithm \ref{alg:2} */
		\STATE $P$ $\gets$ Environmental selection; /* Algorithm \ref{alg:3} */
		\ENDWHILE
		\ENDFOR
		\STATE return $P$;
	\end{algorithmic}
\end{algorithm}

Populations are first obtained by randomly generating different Inception-like neural architectures. These neural architectures are trained to evaluate the individuals in the population efficiently. Then, binary tournament selection is applied to pair individuals within the population. Single-point crossover and period mutation of genes in the mating individuals of the parent population are performed to obtain offspring from the population. After that, the LR schedule obtained by the Meta-LR scheme is used to train the offspring population produced by the parent population, and an adaptive surrogate model is also employed to evaluate the individuals in the offspring population efficiently. Finally, an environment selection based on NSGA-II algorithm is used to select individuals in all populations to retain potentially good individuals. The steps are repeated until the criteria for terminating evolution are met.

\subsection{Meta-LR}\label{section:3.3}
For DNN training, the choice of optimizer plays an important role. Traditional LR schedules are manually designed and inevitably suffer from limited flexibility due to the significant variation in their training dynamics in NAS tasks. Traditional meta-learners are often used as optimizers that utilize gradients as input and produce updating rules. While faster than conventional optimizers are in predefined scenarios, optimizers trained through this series of algorithms usually have difficulty coping with large-scale optimization problems. This is particularly challenging in NAS, where these methods may struggle to generalize over the large number of different neural structures generated during the search process. In contrast to traditional meta-learners, we design a Meta-LR scheme that uses training loss knowledge as input to determine the LR of the parameters. This is because the architectures encoded by the offspring population vary widely, and loss dynamics are more stable and generalizable than gradient statistics are. The LR is then applied when evaluating the architectures for complete epochs, enabling easy transfer to diverse neural architectures searched by the proposed MetaNAS due to the stable optimization process and greatly improving the performance of potential architectures.

In this paper, we choose vanilla SGD algorithm as the optimizer in the search process:
\begin{equation}
	\begin{aligned}	
		\mu_{t+1}=\mu_{t}+\Delta \mu_{t}=\mu_{t}-\alpha_{t} \nabla_{\mu} f_L(D_t;\mu_{t}) \\
		\nabla_{\mu} f_L(D_t;\mu_{t}) = \frac{1}{\lvert D_{t} \rvert} {\textstyle \sum_{i\in D_{t}}}  \nabla_{\mu} f_L(\mu_{t}) 
	\end{aligned}
\end{equation}
where $\mu_{t}$ denotes the $t$ th updating model parameter, $f_L$ represents the training loss function, $\nabla_{\mu} f_L(D_t;\mu_{t})$ denotes the gradient of $f_L$ at $\mu_{t}$, $D_{t}$ represents the batch samples from the training dataset, $i$ represents the data samples in the batch, $\alpha_{t}$ represents current LR at $t$ th iteration and ${\lvert D_{t} \rvert}$ represents the batch size. Typically, the parameter values gradually converge, and the LR decreases during DNN training, however, the change in the decay of the learning rate varies for different LR schedules. Some conventional LR schedules are commonly used in DNNs, such as LambdaLR, MultiStepLR, ExponentialLR, and CosineAnnealingLR. Unfortunately, for complex search process, the predefined LR schedules mentioned above may not always be effective because of the diversity of architectures.

\begin{figure}[htbp]
	\centering
	\includegraphics[scale=1.0, width=\linewidth]{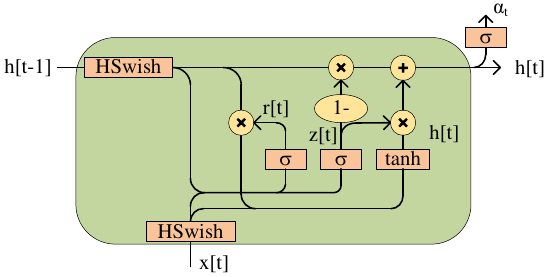}
	\caption{Computational graph of Meta-LR in MetaNAS.}
	\label{fig:2}
\end{figure} 

In response to this issue, we suggest designing an innovative meta-learner with training loss as input when evaluating offspring to obtain Meta-LR. When performing the prediction for complete epochs, Meta-LR can adapt well to the dynamic changes of the search process and provide flexibility for the task. Fig. \ref{fig:2} shows the computational graph of Meta-LR, and the expression of $\alpha_t$ in Eq. (6) can be represented by $G(f_t,h_t;\phi)$:
\begin{equation}	
	\mu_{t+1}=\mu_{t}+\Delta \mu_{t}=\mu_{t}-G(f_t,h_t;\phi) \nabla_{\mu} f_L(D_t;\mu_{t})
\end{equation} 
where $G(f_t,h_t;\phi)$ is the framework of Meta-LR, $\phi$ is the parameter of Meta-LR, $f_t$ denotes the training loss of the current batch $D_t$ at $t$ th iteration, and $h_t$ represents the output at the $t$ th iteration. In each iteration, $G(f_t,h_t;\phi)$ learns the relationship between loss and LR. 

Unlike the vanilla Gate Recurrent Unit, $f_t$ and input $h_t$ are processed with HSwish activation function. Then, it works as a vanilla Gate Recurrent Unit. Moreover, a sigmoid activation function $\sigma$ is employed to derive the predicted value. Since the Meta-LR scheme is parameterized, this simple architecture can effectively capture long-term dependencies in the data with low computational cost. The parameterized structure facilitates the adaptation of architectures encoded by individuals to the correlation between training loss and the learning rate, thereby enabling a rapid response to complex training dynamics. By taking the training loss as input, the meta-learner can learn the LR schedule and generalize to multiple architectures using gradient knowledge of the original problem, which facilitates learning the optimal LR schedule and searching for the best architecture.

\subsection{Adaptive Surrogate Model and Penalty Constraint}\label{section:3.4}
High evaluation cost seriously hinder the design and utilization of NAS methods, leading to the precise selection of individuals with potential for environmental selection. When evaluating offspring individuals, we perform a fast adaptive surrogate evaluation, which can significantly reduce the computational cost of the evolutionary algorithm. We use a simple Radial Basis Function (RBF) surrogate model because of its stable rank correlation coefficient. Kendall's $\tau$ on two datasets is shown in Table \ref{tab:7}, where $\tau$ takes the average of five surrogate evaluations. We rank the performance at the first epoch, rerank the performance after 25 epochs and record Kendall's $\tau$. This coefficient is a widely employed metric of the correlation between two different rankings \cite{sen1968estimates}. $\tau \in [-1, 1]$, where $\tau = 1$ means the two rankings are identical and where $\tau = -1$ means they are completely opposite. Algorithm \ref{alg:2} demonstrates the proposed adaptive surrogate evaluation. 

\begin{table}[t]
	\centering
	\footnotesize
	\renewcommand\arraystretch{1.2}
	\renewcommand\tabcolsep{3.5pt}
	\caption{Kendall's $\tau$ on CIFAR-10 and CIFAR-100.}
	\begin{tabular}{c|c|c}
		\hline
		
		Surrogate & $\tau$ on CIFAR-10 & $\tau$ on CIFAR-100 \\
		\hline
		RBF & 0.80 & 0.76 \\
		MLP & 0.82 & 0.63 \\
		GP & 0.74 & 0.71 \\
		\hline
	\end{tabular}
	\label{tab:7}
\end{table}

\begin{algorithm}[htb]
	\caption{Adaptive surrogate evaluation}
	\label{alg:2}
	\renewcommand{\algorithmicrequire}{ \textbf{Input:}}     
	\renewcommand{\algorithmicensure}{ \textbf{Output:}}    
	\begin{algorithmic}[1] 
		\REQUIRE Population $P$, Training Epochs $M$, Threshold of top1 error $H_t$, offspring $Q$, Complete Training Epochs $M_c$.
		\ENSURE Predicted top1 error $H$.
		
		\STATE Evaluate individuals in $P + Q$ by training architectures encoded by the individuals for $M$ epochs;
		\STATE $S$ $\gets$ Build a surrogate model of individuals in $Q$;
		\STATE $H$ $\gets$ Evaluate top1 error at the point via $S$;
		\IF{$H > H_t$}
		\STATE Solve an auxiliary problem to select the new point to evaluate;
		\STATE $S$ $\gets$ Update $S$;
		\STATE $H_t$ $\gets$ $H$;
		\ELSE
		\STATE Evaluate individuals in $P + Q$ by training architectures encoded by the corresponding individuals with Meta-LR for $M_c$ epochs;
		\STATE $H$ $\gets$ Evaluate top1 error via $S$;
		\ENDIF
		\STATE return $H$;
	\end{algorithmic}
\end{algorithm}

First, a few epochs are evaluated, and the RBF surrogate model is constructed on the basis of the individuals obtained via Algorithm \ref{alg:1}. The surrogate model is subsequently used to obtain the predicted top1 error. When the predicted value reaches the updating threshold, the surrogate model is updated. Otherwise, the offspring are predicted for complete epochs, and the predicted top1 error is returned. By this means, the adaptive surrogate model allows for flexibility in evaluating individuals at different levels. Moreover, the threshold $H_t$ is initially set to 1 and then changed, as shown below.

\begin{equation}
	H_t = 1-\tau\cdot{\rm Acc_{s}}-(1-\tau)\cdot \frac{\rm 1e6}{{Params}}
\end{equation}
where $\tau$ is Kendall's $\tau$, ${\rm Acc_{s}}$ represents the top1 accuracy evaluated by the surrogate model, ${Params}$ represents the number of model parameters, and the scale factor $1e6$ normalizes the number of parameters to a comparable range with accuracy. $\tau$ acts as a dynamic weight to trade off two objectives, and it is updated during evolution process to gradually shift attention from complexity reduction to accuracy maximization. In general, when $\tau$ is increased, the evaluated accuracy ${\rm Acc_{s}}$ increases. Moreover, this may lead to a greater number of model parameters.

Next, the fitness score for each individual is calculated. A dynamic penalty function is employed to penalize the fitness of individuals whose performance does not meet the predefined constraint value $C_{target}$. The dynamic penalty function is represented as follows:
\begin{equation}
	Fitness = \gamma\cdot{\rm Acc_{s}}-(1-\gamma)\cdot \frac{|Complexity-C_{target}|}{|{Complexity_{\operatorname{max}}-C_{target}}|}
\end{equation}
where $Complexity$ represents the complexity of the chosen individual and where $Complexity_{\operatorname{max}}$ represents the maximum complexity of the individuals in the population. $C_{target}$ is the target hardware constraint value. $\gamma$ can be expressed as $\gamma=1-H_t$. In this paper, complexity is defined as the number of parameters in the model. As expected, the underperforming architectures receive lower fitness scores. In the earlier evolution, $\gamma$ is close to 0, and the fitness score is mainly determined by the constraint penalty. As the value of $\gamma$ stabilizes, the influence of accuracy on the fitness score increases significantly.

\begin{figure}[htbp]
	\centering
	\includegraphics[scale=0.6,width=\linewidth]{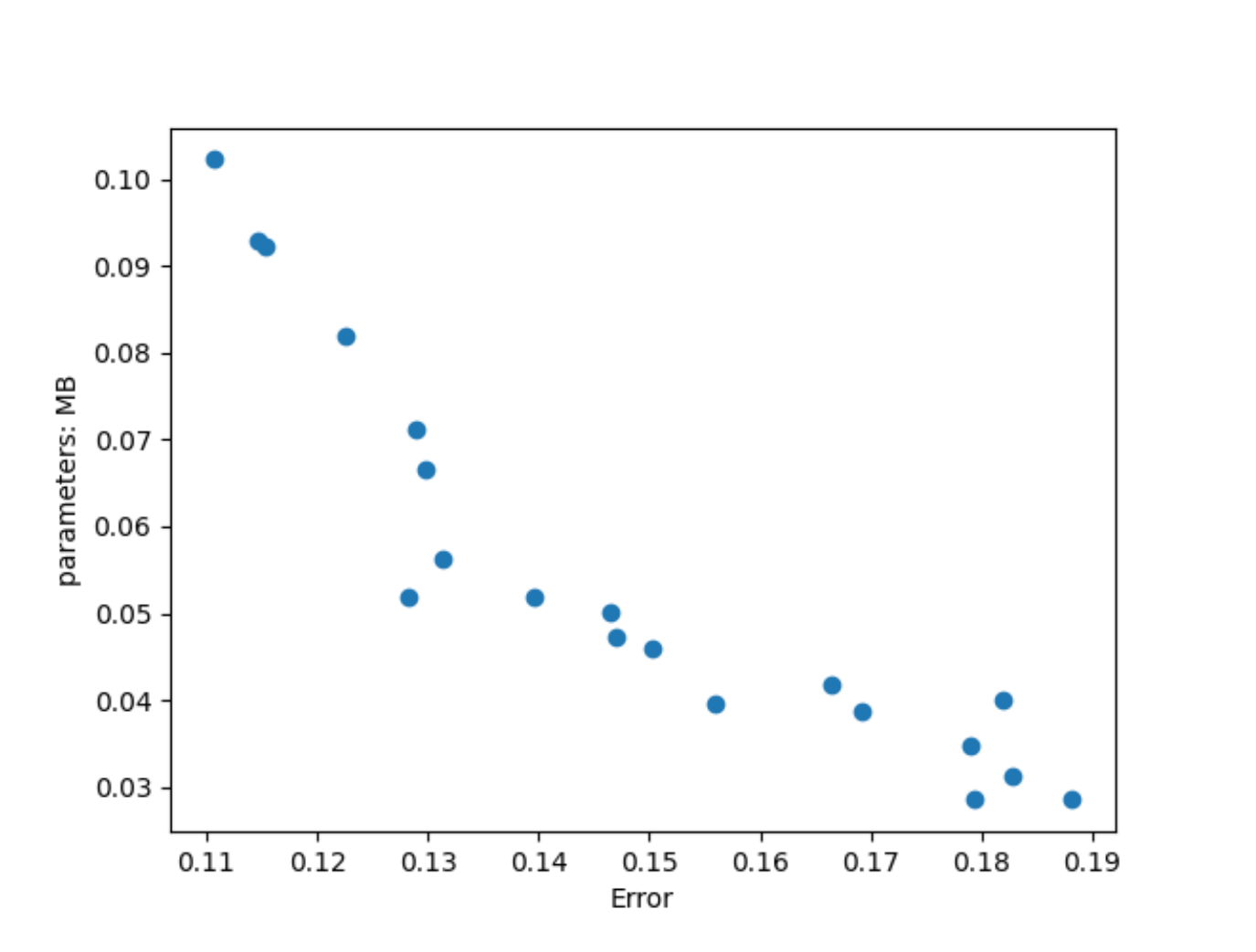}
	\caption{Correlation between the validation error rate and the number of parameters in the architectures on CIFAR-10.}
	\label{fig:3}
\end{figure}

Fig. \ref{fig:3} presents the relationship between the validation error rate and the number of parameters in the architectures implemented by MetaNAS. As the error rate increases, the number of parameters reaches a stable state. Unlike Kendall's $\tau$, which emphasizes only the correlation, the adaptive threshold $H_t$ can help the surrogate model effectively synthesize the number of parameters, evaluate accuracy, and then make a judgment (whether to evaluate for complete epochs).

NSGA-II is a competitive multiobjective EC based on fast nondominated sorting and elitist-preserving strategies \cite{deb2002fast}. The algorithm flow of NSGA-II is described in Algorithm \ref{alg:3}. During the environmental selection process of NSGA-II, $P_s$ promising individuals are selected from $P + Q$ population on the basis of the crowding distance and elitist nondominated sorting.

\begin{algorithm}[htb]
	\caption{NSGA-II Environmental Evaluation}
	\label{alg:3}
	\renewcommand{\algorithmicrequire}{ \textbf{Input:}}     
	\renewcommand{\algorithmicensure}{ \textbf{Output:}}    
	\begin{algorithmic}[1] 
		\REQUIRE Population $P$, Population Size $P_s$, offspring $Q$, total number of iterations $T$.
		\ENSURE Population $P$.
		
		\STATE Initialize the index $t=1$;
		\STATE Initialize the population $P(1)$ with $P_s$;
		\STATE Calculate the objective values for $P(1)$;
		\FOR {index $t=1,2,...,T$}
		\STATE Generate $P_s$ offspring $Q$ by applying genetic operators on $P(t)$;
		\STATE Calculate objective values of $Q(t)$;
		\STATE $R(t)$ $\gets$ Combine $P(t) + Q(t)$;
		\STATE Sort $R(t)$ based on dominance relationship and crowding distance;
		\STATE Select the best $P$ individuals from $R(t)$ as new population $P(t + 1)$
		\STATE $t$ $\gets$ $t+1$;
		\ENDFOR
		\STATE return $P$;
	\end{algorithmic}
\end{algorithm}

The main idea of nondominated sorting is to generate a series of ordered fronts on the basis of the Pareto dominance relationship between the objective values of $R(t)$ and solutions within the same front that cannot dominate each other. Solutions in the first nondominated fronts are given higher priority to be selected. The sorting algorithm has a computational complexity of $O(n{P_s}^2)$, where $n$ is the number of objectives and $P_s$ is the population size. A crowding distance that measures the distance between two neighboring solutions in the same front is calculated to promote solution diversity, and those having a large crowding distance are prioritized for selection. The computational complexity of crowding distance calculation is $O(n{P_s}^2log{P_s})$ in the worst case, when all the solutions are located in one nondominated front.

\subsection{Genetic Operator}\label{section:3.7}
A tailored genetic operator that incorporates crossover and mutation is designed for the proposed MetaNAS. A detailed explanation is given below.

The crossover operator is a two-part operator: inter-crossover and intra-crossover. Inter-crossover refers to the swap of normal cells and reduction cells between two parent individuals. Intra-crossover employs single-point crossover, exchanging links and operations in the normal cells and reduction cells of the two parent individuals. For example, we set two normal cells $C_0$ and $C_1$.
\begin{equation}
	\begin{aligned}	
		C_0 &= (\bold{n_0^0}, \bold{n_1^0}, ...,\bold{n_j^0}, \bold{n_{N_0-1}^0}) \\
		C_1 &= (\bold{n_0^1}, \bold{n_1^1}, ...,\bold{n_j^1}, \bold{n_{N_1-1}^1})
	\end{aligned}
\end{equation}

where $\bold{n_j^0}$ and $\bold{n_j^1}$ represent the subvectors of $C_0$ and $C_1$, respectively. Suppose that $N_0 \le N_1$; when a random number $Rand$ is sampled from the vector interval $[0,N_1]$ that satisfies $Rand < N_0$, offspring normal cells $C^{'}_0$ and $C^{'}_1$ can be represented as follows:
\begin{equation}
	\begin{aligned}	
		C^{'}_0 &= C_1[0:Rand] + C_0[Rand+1:N_0] \\
		C^{'}_1 &= C_0[0:Rand] + C_1[Rand+1:N_1]
	\end{aligned}
\end{equation}
where $Rand < N_0$ ensures that the crossover point is located in the shorter parent.

The mutation probabilities for links and operations in cells are not identical. The probability of link mutation is related to the length of the vector representing the architecture and index of operations, whereas the probability of operation mutation is related only to the index of operations. In addition to the flip bit mutation, another mutation component involves adding a randomly generated extra node or removing a node. In particular, to be consistent with the current trend of accelerated evolution, we design the period mutation operator to find more diverse and potential architectures and underscore the enhanced generalizability of the optimal architecture.
\begin{equation}
	[M_{l},M_{o}] =
	\begin{cases}
		\displaystyle[\frac{1}{l_v-l_o},\frac{1}{l_o}] , 	\quad\quad \left [kN_l, kN_l+N_h \right ), k = 1,2,... \\ 
		\displaystyle[1-\frac{1}{l_v-l_o},1-\frac{1}{l_o}] , \quad others
	\end{cases}
\end{equation}
where $M_{l}$ and $M_{o}$ are the probabilities of link mutation and operation mutation, respectively. $l_v$ represents the length of the vector, and $l_o$ represents the index of operations. The mutation probability briefly increases for every $N_l$ individuals and returns to the original probability after $N_h$ individuals.

\subsection{Method Details}
1) \textbf{Population Initialization.} 
On the basis of the proposed encoding strategy, we introduce the details of population initialization. The set of randomly generated Inception-like architectures can be obtained by initializing the links and operations as follows: the first two bits of links in each node are set to 1 with a high probability, whereas the remaining bits are randomized to 0 or 1. The operations for each node are selected randomly from the different operations listed in Table \ref {tab:1}. \\

2) \textbf{Encoding Strategy.}
Each Inception-like architecture is encoded by an individual $P_j$, including two vectors to represent normal cell and reduction cell. The cells consist of multiple nodes, with each node representing a subvector. For example, the subvector $\bold{n_j}$ can be expressed as follows:
\begin{equation}
	\begin{aligned}	
		\bold{n_j} &= (Link,Operation)\\
		&=({l_{in_0}}, {l_{in_1}}, l_0, l_1,...,l_t,...,l_{j-1},Operation), \\
		&t=0,1,...,j-1, l_t\in{\{0,1\}}, Operation\in{\{1,2,...,12\}}
	\end{aligned}
\end{equation}
$Link=({l_{in_0}}, {l_{in_1}}, l_0, l_1,...,l_t,...,l_{j-1})$ is in node $j$, where ${l_{in_0}}$ and ${l_{in_1}}$ denote the links to the input nodes ${in_0}$ and ${in_1}$, respectively. $l_t$ represents the link between nodes $j$ and node $t$. $Operation$ corresponds to the operations listed in Table \ref {tab:1}. \\

\begin{figure}[htbp]
	\centering
	\includegraphics[scale=0.7,trim=0 0 0 0]{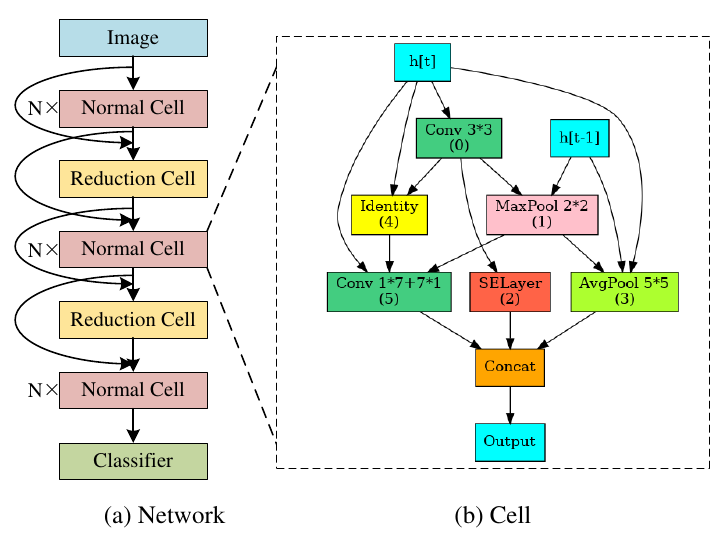}
	\caption{Search space of the proposed MetaNAS.}
	\label{fig:4}
\end{figure}

3) \textbf{Search Space.}
Fig. \ref{fig:4}(a) depicts that the neural network is constructed by a cell-based search space.

These two kinds of cells (normal cells and reduction cells) take the outputs of two previous cells as inputs. Each cell can be defined as a convolutional network that maps two $H \times W \times F$ tensors to one $H{'} \times W{'} \times F{'}$ tensor. As described in \cite{elsken2019neural}, operations with stride 1 are set for normal cells, and reduction cells apply the same operations with a stride of 2, where $H{'} = H/2$, $W{'} = W/2$ and $F{'} = 2F$. Directed acyclic graphs (DAGs) formed by nodes are used to construct these two types of cells. In Fig. \ref{fig:4}(b), an example cell designed by the proposed algorithm is shown. This cell consists of two input nodes, one output node and several hidden nodes. 

\begin{table}[htb]
	\centering
	\footnotesize
	\renewcommand\arraystretch{1.2}
	\renewcommand\tabcolsep{3.5pt}
	\caption{All the operations in the search space.}
	\begin{tabular}{c|c|c}
		\hline
		
		Index&Operation&Kernel Size \\
		\hline
		1 & Identity mapping & - \\
		2 & Conv 1*1 & 1*1 \\
		3 & Conv 3*3 & 3*3 \\
		4 & Conv 1*3+3*1 & 1*3+3*1 \\
		5 & Conv 1*7+7*1 & 1*7+7*1 \\
		6 & MaxPool 2*2 & 2*2 \\
		7 & MaxPool 3*3 & 3*3 \\
		8 & MaxPool 5*5 & 5*5 \\
		9 & AvgPool 2*2 & 2*2 \\
		10 & AvgPool 3*3 & 3*3 \\
		11 & AvgPool 5*5 & 5*5 \\
		12 & SELayer & - \\
		\hline
	\end{tabular}
	\label{tab:1}
\end{table}

The basic components for constructing cells are nodes. For each node, inputs are combined (summed or concatenated) as outputs after their respective operations. The structure of a DAG depends on different links and operations of the inner nodes. Table \ref {tab:1} shows all the predefined computation operations.

\section{Experimental}\label{section:4}
\label{sec: 4}
In this section, we first introduce the related experimental settings of our proposed MetaNAS. Several experiments comparing MetaNAS and several state-of-the-art methods prove the superiority of the proposed method. Finally, we analyze the proposed Meta-LR scheme and adaptive surrogate model in detail.
\subsection{Datasets}\label{section:4.1}
The experiments utilize three commonly used benchmark datasets comprising natural images: CIFAR-10, CIFAR-100 \cite{krizhevsky2009learning}, and ImageNet1K \cite{deng2009imagenet}. Both CIFAR-10 and CIFAR-100 datasets consist of 60K RGB images with a spatial resolution of 32 $\times$ 32, including 50k training images and 10k testing images. Horizontal flipping with 0.5 probability is applied to all training images during the search and training process, and cutout augmentation is used in the training process. ImageNet1K consists of about 1.28M high-resolution images belonging to 1,000 different categories. To ensure a fair comparison, the input size is standardized to 224 $\times$ 224. In addition, we utilize some commonly used augmentation techniques, including random resized crop, horizontal flip and color jitter.
\subsection{Parameter settings}\label{section:4.2}
MetaNAS uses the following settings:

\textbf{Details in meta-learning.} We compare Meta-LR with commonly used LR schedules: exponential decay momentum-SGD, cosine annealing momentum-SGD, and adaptive LR schedule Adam. We use MobileNetV2 to train on CIFAR-10 and CIFAR-100, with batch size 128. The initial LR is 0.05 for exponential decay momentum-SGD and cosine annealing momentum-SGD, and the momentum coefficient is 0.9. Exponential decay momentum-SGD has a decay rate of 0.97 every epoch. Cosine annealing momentum-SGD decays to 0.001, and the weight decay is $5 \times 10^{-4}$. The Adam optimizer has a learning rate of 0.01, a momentum of (0.5, 0.999), and a weight decay of $5 \times 10^{-4}$. The initial LR in Meta-LR is $1 \times 10^{-3}$, and the weight decay is $1 \times 10^{-4}$. Both CIFAR-10 and CIFAR-100 are trained with ResNet-18 for 100 epochs, and the LR is continuously updated with the current training loss using Meta-LR.

\textbf{Details in search process.} In each cell, the number of nodes ranges from 5--12, and every network comprises 20 cells. For normal cells, the stride for operations is set to 1, whereas reduction cells utilize operations with a stride of 2. The channel number in each individual is set to 40, population size $P_s$ is set to 20, maximum number of evolutionary generations $G = 5$, and number of training epochs $M = 35$. The number of complete training epochs $M_c = 100$ on CIFAR-10, where the LR uses the LR schedule obtained by the Meta-LR scheme in pretraining. The search process is executed on one NVIDIA GeForce RTX 3090 GPU. As shown in Fig. \ref{fig:3}, the quantity of parameters is relatively stable when the error rate reaches a certain value on CIFAR-10. The mutation period $N_l$ is 4, and $N_h$ is 1. 

\textbf{Details in training process.} The batch size is set to 128 for CIFAR-10 and CIFAR-100 and 512 for ImageNet1K. All training images undergo standard augmentation. Specifically, we randomly crop $32\times32$ patches from upsampled images of size $40\times40$ and apply random horizontal flips with a probability of 0.5. Additionally, cutout augmentation is applied during the training process. The Pareto optimal individuals are obtained from the search process. Some potential architectures encoded by these individuals are trained on CIFAR-10, CIFAR-100 and ImageNet1K. The training process is also executed on one NVIDIA GeForce RTX 3090 GPU.

1) On CIFAR-10 and CIFAR-100, we train the optimal architecture for 600 epochs using momentum-SGD optimizer. The following settings are implemented: initial learning rate of 0.025, momentum coefficient of 0.9, batch size of 128, L2 weight decay of $5 \times 10^{-4}$ and dropout rate of 0.4 applied in the final layer. Furthermore, each path is abandoned with a probability of 0.2. 

2) On ImageNet1K, the optimal architecture is trained for 250 epochs using momentum-SGD optimizer with momentum coefficient of 0.9 and batch size of 256. The initial learning rate is set to 0.1, which decays by a factor of 0.97 per epoch. The rest of the settings are the same as those of CIFAR-10 and CIFAR-100.

\begin{table}[htb]
	\begin{threeparttable}
		\centering
		\footnotesize
		\renewcommand\arraystretch{1.2}
		\renewcommand\tabcolsep{3.5pt}
		\caption{Comparison of MetaNAS with state-of-the-art methods on CIFAR-10}
		\begin{tabular}{ccccc}
			\hline
			
			Architecture&Method&Top1 acc(\%)&Params(M)&GPU days\\
			\hline
			VGG \cite{simonyan2014very} & manual & 93.34 & 138.36 & - \\
			Resnet-101 \cite{he2016deep} & manual & 93.57 & 44.55 & - \\
			MobileNetV2 \cite{sandler2018mobilenetv2} & manual & 94.56 & 3.51 & - \\
			\hline
			NASNet-A \cite{zoph2018learning} & RL & 97.35 & 3.3 & 2000 \\
			MetaQNN \cite{baker2016designing} & RL & 93.08 & 11.2 & 80 \\
			DARTS \cite{liu2018darts} & GD & 97.00 & 3.3 & 1.5 \\
			NAO \cite{luo2018neural} & GD & 96.82 & 10.6 & 200 \\
			DSO-NAS-share \cite{zhang2020you} & GD & 97.26 & 3.0 & 1.0 \\
			\hline
			Genetic-CNN \cite{xie2017genetic} & EC & 92.90 & - & 17 \\
			Large-scale Evolution \cite{real2017large} & EC & 94.60 & 5.4 & 2750 \\
			Amoebanet-A \cite{real2019regularized} & EC & 96.66 & 3.2 & 3150 \\
			CNN-GA \cite{sun2020automatically} & EC & 95.22 & 2.9 & 35 \\
			NSGANet \cite{lu2019nsga} & EC & 97.25 & 3.3 & 4 \\
			\underline{NSGANetV2-s}\tnote{*} \cite{lu2020nsganetv2} & EC & 98.40 & 6.1 & 2.2 \\
			AE-CNN \cite{sun2020completely} & EC & 95.70 & 2.0 & 27 \\
			AE-CNN + E2EPP \cite{sun2019surrogate} & EC & 94.70 & 7 & 4.3 \\
			CARS-E \cite{yang2020cars} & EC & 97.14 & 3.0 & 0.4 \\
			NPENAS-BO \cite{wei2022npenas} & EC & 97.46 & 3.5 & 1.8 \\
			Evo-OSNet \cite{zhang2023evolutionary} & EC & 97.44 & 3.3 & 0.5 \\
			SI-EvoNet-S \cite{zhang2020efficient} & EC & 97.31 & 1.84 & 0.458 \\
			EPCNAS-C \cite{huang2022particle} & EC & 96.93 & 1.16 & 1.10 \\
			MOEA-PS \cite{xue2023neural} & EC & 97.23 & 3.0 & 2.6 \\
			PVLL-NAS++ \cite{li2022neural} & EC & 97.25 & 3.23 & 0.22 \\
			\textbf{MetaNAS} & EC & \textbf{97.88 $\pm$ 0.08} & 2.91 & 0.833 \\
			\hline
		\end{tabular}
		\label{tab:2}
		\begin{tablenotes}
			\item[*] The search cost excludes the supernet training cost.
		\end{tablenotes}
	\end{threeparttable}
\end{table}

\begin{table}[htb]
	\centering
	\footnotesize
	\renewcommand\arraystretch{1.2}
	\renewcommand\tabcolsep{3.5pt}
	\caption{Comparison of MetaNAS with state-of-the-art methods on CIFAR-100}
	\begin{tabular}{ccccc}
		\hline
		
		Architecture&Method&Top1 acc(\%)&Params(M)&GPU days\\
		\hline
		VGG \cite{simonyan2014very} & manual & 67.95 & 138.36 & - \\
		Resnet-101 \cite{he2016deep} & manual & 74.84 & 44.55 & - \\
		MobileNetV2 \cite{sandler2018mobilenetv2} & manual & 77.09 & 3.51 & - \\
		\hline
		NASNet-A \cite{zoph2018learning} & RL & 83.42 & 3.3 & 2000 \\
		MetaQNN \cite{baker2016designing} & RL & 72.86 & 11.2 & 80 \\
		DARTS \cite{liu2018darts} & GD & 82.46 & 3.4 & 1.5 \\
		NAO \cite{luo2018neural} & GD & 84.33 & 10.8 & 200 \\
		\hline
		Genetic-CNN \cite{xie2017genetic} & EC & 70.97 & - & 17 \\
		Large-scale Evolution \cite{real2017large} & EC & 77.00 & 40.4 & 2750 \\
		Amoebanet-B \cite{real2019regularized} & EC & 84.20 & 3.2 & 3150 \\
		CNN-GA \cite{sun2020automatically} & EC & 79.47 & 4.1 & 40 \\
		NSGANet \cite{lu2019nsga} & EC & 79.26 & 3.3 & 4 \\
		AE-CNN \cite{sun2020completely} & EC & 79.15 & 5.4 & 36 \\
		AE-CNN + E2EPP \cite{sun2019surrogate} & EC & 77.98 & 20.9 & 10 \\
		Evo-OSNet \cite{zhang2023evolutionary} & EC & 84.16 & 3.32 & 0.8 \\
		SI-EvoNet-S \cite{zhang2020efficient} & EC & 84.30 & 3.32 & 0.813 \\
		EPCNAS-C \cite{huang2022particle} & EC & 81.67 & 1.29 & 1.10 \\
		MOEA-PS \cite{xue2023neural} & EC & 81.03 & 5.8 & 5.2 \\
		\textbf{MetaNAS} & EC & \textbf{84.49 $\pm$ 0.06} & 2.96 & 0.833 \\
		\hline
	\end{tabular}
	\label{tab:3}
\end{table}

\subsection{Result and Analysis}\label{section:4.3}
According to the parameter settings, we obtain the experimental results of the optimal neural network on CIFAR-10 and CIFAR-100. A comparison of state-of-the-art methods is shown in Table \ref {tab:2} and Table \ref {tab:3}.

In Table \ref {tab:2} and Table \ref{tab:3}, the optimal network searched by our proposed method achieves 97.88\% accuracy on CIFAR-10 and 84.49\% accuracy on CIFAR-100 with fewer than 3M parameters. NSGANetV2-s takes only 2.2 GPU days to reach 98.4\% accuracy because it does not include the search cost on ImageNet1K. The state-of-the-art methods for comparison contain manually designed network structures, and some non-EC-based NAS methods include RL-based methods and gradient-based methods. In addition, many EC-based NAS methods are compared.

To confirm the effectiveness of our proposed MetaNAS on complex datasets, we transferred the optimal architecture searched on CIFAR-10 to ImageNet1K for further relevant experiments as well as a comparison of the results, as shown in Table \ref {tab:4}.

\begin{table}[htb]
	\centering
	\footnotesize
	\renewcommand\arraystretch{1.2}
	\renewcommand\tabcolsep{3.5pt}
	\caption{Comparison between the optimal architecture on CIFAR-10 transfer to ImageNet1K and other methods}
	\begin{threeparttable}
	\begin{tabular}{cccccc}
		\hline
		\multirow{2}{*}{Architecture}&\multirow{2}{*}{Method}&\multicolumn{2}{c}{Test accuracy}&\multirow{2}{*}{Params(M)}&\multirow{2}{*}{GPU days} \\
		\cline{3-4}
		~ & ~ & Top1(\%)&Top5(\%) ~ & ~ \\
		\hline
		InceptionV1 \cite{szegedy2015going} & manual & 69.8 & 89.9 & 6.6 & - \\
		ShuffleNet \cite{zhang2018shufflenet} & manual & 73.6 & 89.8 & 5 & - \\
		MobileNetV2 \cite{sandler2018mobilenetv2} & manual & 70.6 & 89.5 & 4.2 & - \\
		\hline
		NASNet-A \cite{zoph2018learning} & RL & 74.0 & 91.6 & 5.3 & 2000 \\
		DARTS \cite{liu2018darts} & GD & 73.3 & 91.3 & 4.7 & 4 \\
		NAO \cite{luo2018neural} & GD & 74.3 & 91.8 & 11.4 & 200 \\
		\hline
		Genetic-CNN \cite{xie2017genetic} & EC & 72.13 & 90.26 & 156 & 17 \\
		Amoebanet-B \cite{real2019regularized} & EC & 74.00 & 91.50 & 5.3 & 3150 \\
		\underline{NSGANetV2-s}\tnote{*}  \cite{lu2020nsganetv2} & EC & 77.40 & 93.50 & 6.1 & 10 \\
		\underline{Evo-OSNet}\tnote{*}  \cite{zhang2023evolutionary} & EC & 77.48 & 93.53 & 5.0 & 8.6 \\
		SI-EvoNet-S \cite{zhang2020efficient} & EC & 75.83 & 92.59 & 4.7 & 0.458 \\
		MOEA-PS \cite{xue2023neural} & EC & 73.60 & 91.50 & 4.7 & 2.6 \\
		CARS-E \cite{yang2020cars} & EC & 73.70 & 91.60 & 4.4 & 0.4 \\
		\underline{CENAS-A}\tnote{*}  \cite{ma2023pareto} & EC & 77.40 & 92.80 & 3.24 & 1.91 \\
		\textbf{MetaNAS} & EC & \textbf{76.42} & 92.67 & 2.91 & 0.833 \\
		\hline
	\end{tabular}
	\label{tab:4}
	\begin{tablenotes}
		\item[*] The methods are directly searched on ImageNet1K.
	\end{tablenotes}
	\end{threeparttable}
\end{table}

\begin{figure}[htbp]
	\centering
	\includegraphics[scale=0.3,trim=0 0 0 0]{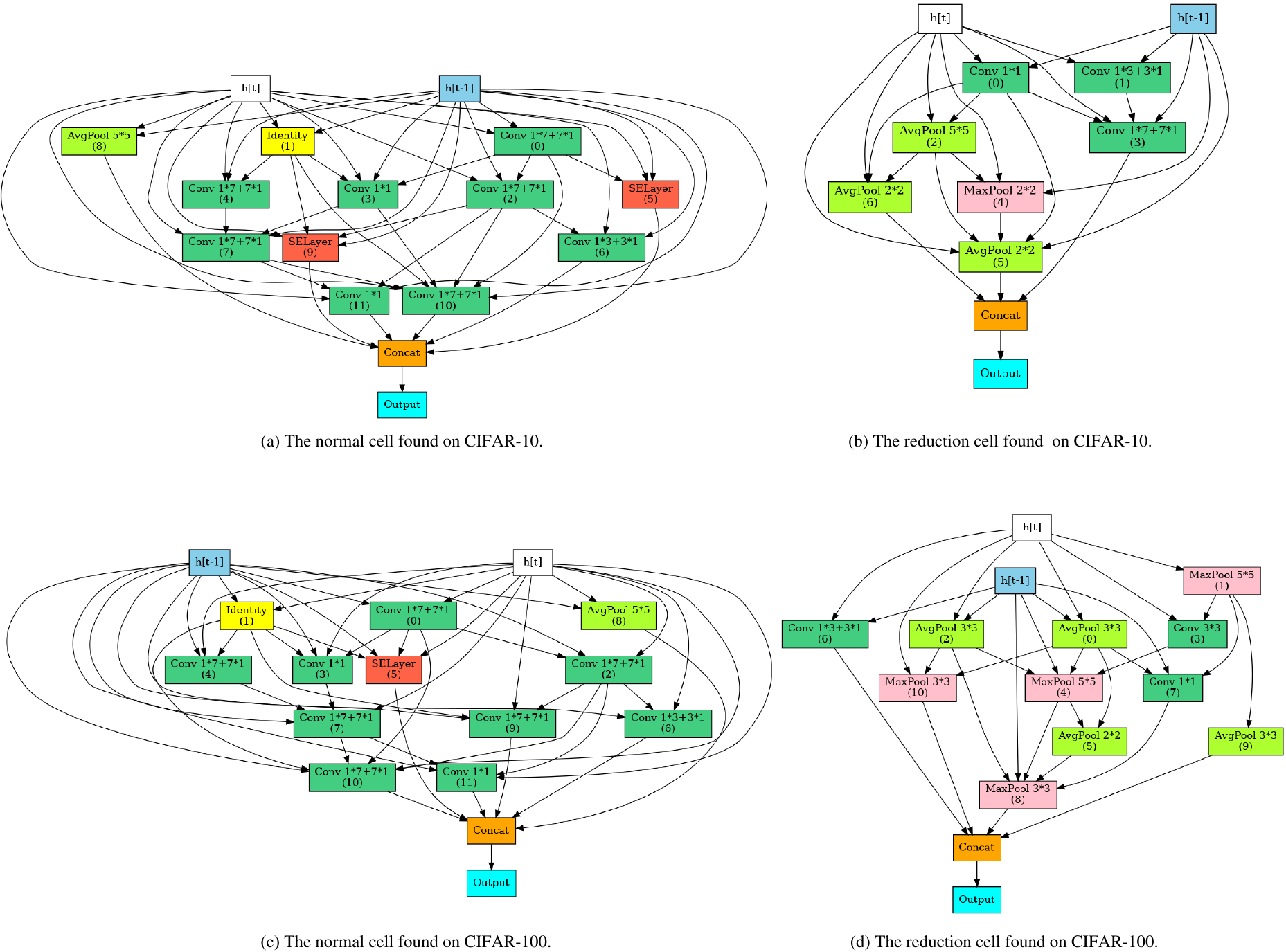}
	\caption{Optimal architectures obtained by MetaNAS on CIFAR-10 and CIFAR-100.}
	\label{fig:5}
\end{figure}

Noticeably, MetaNAS achieves excellent performance while keeping the search cost low compared with other state-of-the-art methods. CENAS-A, Evo-OSNet and NSGANetV2-s can achieve accuracies of 77.40\%, 77.48\% and 77.40\%, respectively. However, their architectures are directly searched on ImageNet1K. For the remaining methods, our MetaNAS achieves competitive performance. The global optimal network architectures searched by MetaNAS on CIFAR-10 and CIFAR-100 are shown in Fig. \ref{fig:5}. 

\subsection{Ablation study}\label{section:4.4}
The ablation experiments on CIFAR-10 and CIFAR-100 for the period mutation operator, adaptive surrogate model and Meta-LR scheme are shown in Table \ref{tab:5} and Table \ref{tab:6}, respectively.
\begin{table}[htb]
	\centering
	\footnotesize
	\renewcommand\arraystretch{1.2}
	\renewcommand\tabcolsep{3.5pt}
	\caption{Ablation study results for Meta-LR, the surrogate model and period mutation on CIFAR-10}
	\begin{tabular}{cccccc}
		\hline
		\multicolumn{3}{c}{Components}&\multirow{2}{*}{Method}&\multirow{2}{*}{Top1 acc(\%)}&\multirow{2}{*}{GPU days} \\
		\cline{1-3}
		Meta-LR & Surrogate & Period Mutation & ~ & ~ & ~ \\
		\hline
		~ & ~ & ~ & EC & 95.70 $\pm$ 0.10 & 2.2 \\
		\checkmark & ~ & ~ & EC & 97.30 $\pm$ 0.05 & 2.5 \\
		~ & \checkmark & ~ & EC & 96.63 $\pm$ 0.08 & 1.0 \\
		~ & ~ & \checkmark & EC & 96.91 $\pm$ 0.12 & 2.2 \\
		\hline
		\checkmark & \checkmark & ~ & EC & 97.60 $\pm$ 0.07 & 0.833 \\
		\checkmark & ~ & \checkmark & EC & 97.51 $\pm$ 0.10 & 2.5 \\
		~ & \checkmark & \checkmark & EC & 97.12 $\pm$ 0.14 & 0.75 \\
		\hline
		\checkmark & \checkmark & \checkmark & EC & 97.88 $\pm$ 0.08 & 0.833 \\
		\hline
	\end{tabular}
	\label{tab:5}
\end{table}
\begin{table}[htb]
	\centering
	\footnotesize
	\renewcommand\arraystretch{1.2}
	\renewcommand\tabcolsep{3.5pt}
	\caption{Ablation study results for Meta-LR, the surrogate model and period mutation on CIFAR-100}
	\begin{tabular}{cccccc}
		\hline
		\multicolumn{3}{c}{Components}&\multirow{2}{*}{Method}&\multirow{2}{*}{Top1 acc(\%)}&\multirow{2}{*}{GPU days} \\
		\cline{1-3}
		Meta-LR & Surrogate & Period Mutation & ~ & ~ & ~ \\
		\hline
		~ & ~ & ~ & EC & 79.42 $\pm$ 0.07 & 2.2 \\
		\checkmark & ~ & ~ & EC & 82.66 $\pm$ 0.05 & 2.5 \\
		~ & \checkmark & ~ & EC & 81.15 $\pm$ 0.04 & 1.0 \\
		~ & ~ & \checkmark & EC & 81.58 $\pm$ 0.07 & 2.2 \\
		\hline
		\checkmark & \checkmark & ~ & EC & 84.03 $\pm$ 0.04 & 0.833 \\
		\checkmark & ~ & \checkmark & EC & 84.23 $\pm$ 0.06 & 2.5 \\
		~ & \checkmark & \checkmark & EC & 83.87 $\pm$ 0.09 & 0.75 \\
		\hline
		\checkmark & \checkmark & \checkmark & EC & 84.49 $\pm$ 0.06 & 0.833 \\
		\hline
	\end{tabular}
	\label{tab:6}
\end{table}
As shown in Table \ref{tab:5} and Table \ref{tab:6}, the Meta-LR scheme is used to increase the robustness of architectures at the end of training, which can help to search for more efficient architectures. Period mutation + Meta-LR amplifies the impact of the Meta-LR generalizability through architectural diversity, leading to high classification accuracy. The adaptive surrogate model guarantees the performance of the final architecture and introduces an adaptive threshold to further reduce the computational cost. Notably, the adaptive threshold itself does not consume computational resources.

\begin{table}[htb]
	\centering
	\footnotesize
	\renewcommand\arraystretch{1.2}
	\renewcommand\tabcolsep{3.5pt}
	\caption{Comparison of different genetic operators on CIFAR-10 and CIFAR-100}
	\begin{tabular}{ccc}
		\hline
		Genetic Operators & CIFAR-10(\%) & CIFAR-100(\%) \\
		\hline
		Mutation probability 0.1 & 95.45 $\pm$ 0.13 & 80.90 $\pm$ 0.11 \\
		Mutation probability 1 & 97.02 $\pm$ 0.08 & 82.81 $\pm$ 0.06 \\
		Period Mutation & 96.99 $\pm$ 0.08 & 82.95 $\pm$ 0.07 \\
		Crossover + Mutation probability 0.1 & 97.51 $\pm$ 0.08 & 84.24 $\pm$ 0.06 \\
		Crossover + Mutation probability 1 & 97.24 $\pm$ 0.10 & 83.76 $\pm$ 0.07 \\
		Crossover + Period Mutation & 97.88 $\pm$ 0.08 & 84.49 $\pm$ 0.06 \\
		\hline
	\end{tabular}
	\label{tab:8}
\end{table}
In Table \ref{tab:8}, we discuss different genetic operators in the proposed MetaNAS. The crossover mentioned stands for single-point crossover. Compared with mutation alone, single-point crossover + mutation can effectively improve model performance. When the mutation probability is 0.1, single-point crossover works much better. The period mutation can achieve higher accuracy with single-point crossover than with only period mutation because the population evolves at a slow rate when only a low mutation probability is used. 

\begin{table}[htb]
	\centering
	\footnotesize
	\renewcommand\arraystretch{1.2}
	\renewcommand\tabcolsep{3.5pt}
	\caption{Ablation study of varying mutation periods on CIFAR-10}
	\begin{tabular}{ccc}
		\hline
		Period $N_l$ & Period $N_h$ & Accuracy(\%) \\
		\hline
		2 & 1 & 97.62 $\pm$ 0.09 \\
		4 & 1 & 97.88 $\pm$ 0.08 \\
		6 & 1 & 97.65 $\pm$ 0.08 \\
		2 & 2 & 97.33 $\pm$ 0.12 \\
		\hline
	\end{tabular}
	\label{tab:9}
\end{table}
As shown in Table \ref{tab:9}, the optimal configuration is $N_l=4$ and $N_h=1$, achieving the highest mean accuracy of 97.88\% $\pm$ 0.08\% on CIFAR-10, suggesting that moderate mutation intervals balance exploration and stability. When $N_l$ is too large, the accuracy decreases. Additionally, increasing $N_h$ introduces instability and degrades model performance.

\begin{figure}[htb]
	\centering
	\includegraphics[scale=0.32,trim=0 0 0 0]{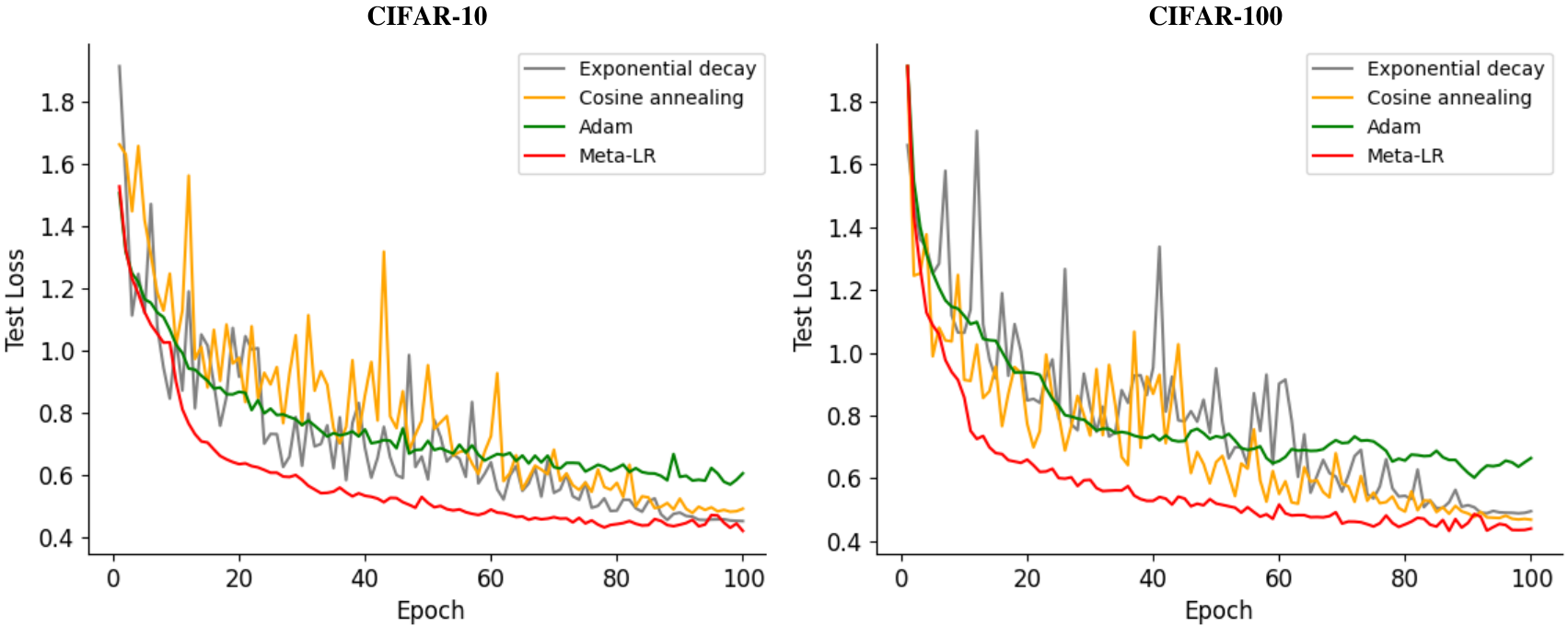}
	\caption{Comparing the effectiveness of the proposed Meta-LR and other LR schedules on three benchmark datasets.}
	\label{fig:6}
\end{figure}

To analyze the performance of the Meta-LR scheme, Fig. \ref{fig:6} shows the test loss on CIFAR-10 and CIFAR-100 for 100 epochs with different LR schedules during the evaluation. 
The Meta-LR scheme is stable and faster in achieving lower test loss than other commonly used LR schedules are. The exponential decay momentum-SGD and cosine annealing momentum-SGD can ultimately reach losses similar to those of other schedules, but more training time is needed. The traditional adaptive LR schedule Adam behaves poorly, therefore, Adam needs to make more adjustments in response to NAS tasks.

\begin{figure}[htb]
	\centering
	\includegraphics[width=1.0\linewidth]{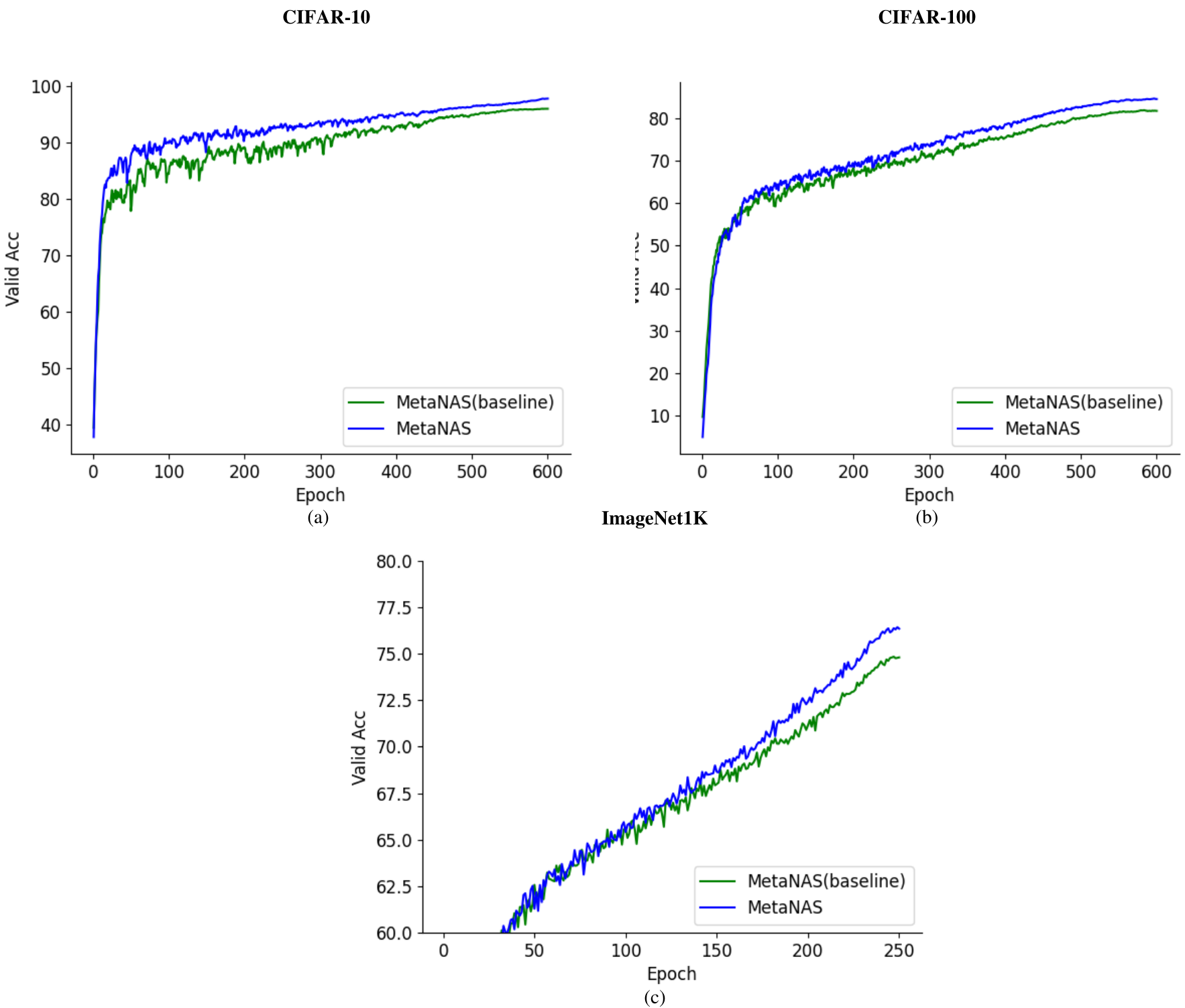}
	\caption{Comparing the effectiveness of the proposed Meta-LR + adaptive surrogate + period mutation on three benchmark datasets.}
	\label{fig:7}
\end{figure}

The training process for different architectures on CIFAR-10, CIFAR-100 and ImageNet1K are depicted in Fig. \ref{fig:7}. MetaNAS (baseline) refers to our method without Meta-LR, surrogate model or period mutation. In Fig. \ref{fig:7}(a), the valid accuracy becomes stable for MetaNAS after 200 training epochs, whereas the baseline takes 300 epochs to achieve a similar level of stabilization. The model stability and performance advantages are further emphasized on CIFAR-100 in Fig. \ref{fig:7}(b). Moreover, the performance of MetaNAS is superior to that of the baseline throughout the training process. As Fig. \ref{fig:7}(c) shows, the model transferred from CIFAR-10 shows competitive performance on ImageNet1K.

 \begin{figure}[htbp]
	\centering
	\includegraphics[scale=0.28,trim=0 0 0 0]{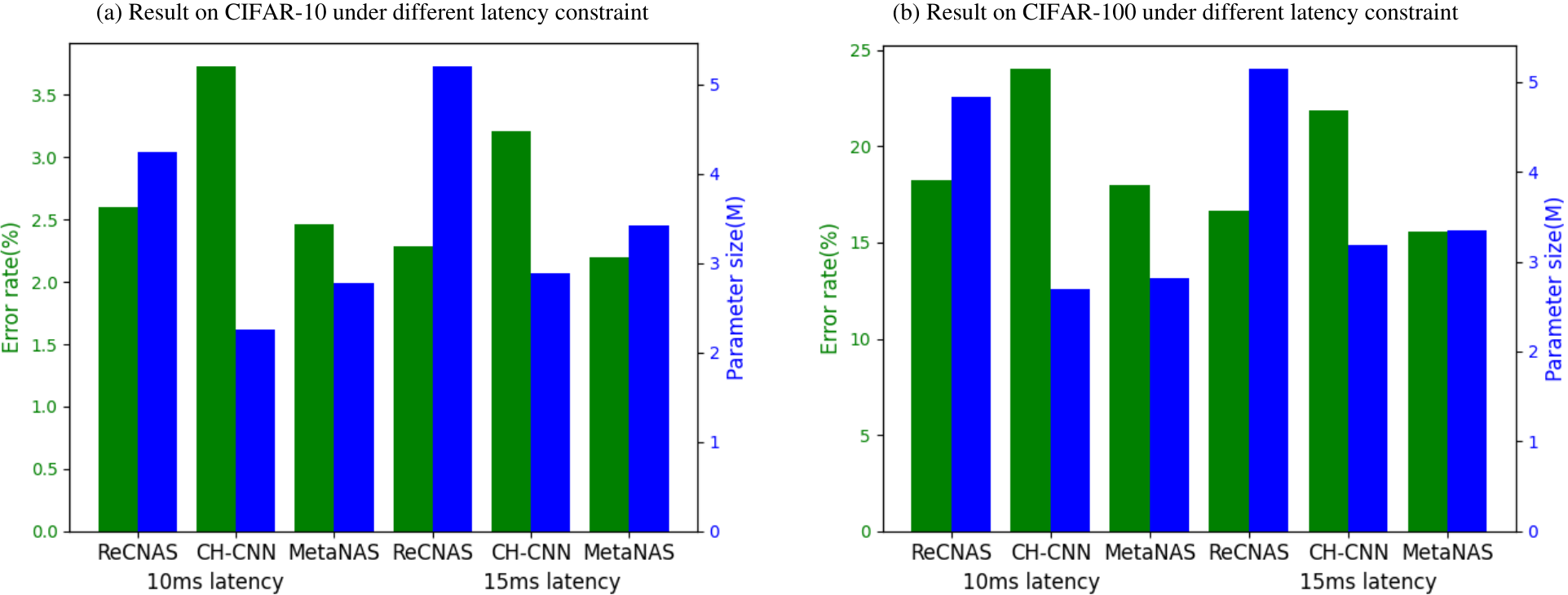}
	\caption{Comparison of search costs under different constraints.}
	\label{fig:8}
\end{figure} 

\subsection{Robustness}\label{section:4.5}
To explore the performance of the proposed method under hardware constraints, we compare it with ReCNAS \cite{peng2024recnas} (GD-based) and CH-CNN \cite{li2021automatic} (EC-based) at different inference latencies.
As Fig. \ref{fig:8} shown, the model obtained by CH-CNN does not perform as well as the proposed method does. MetaNAS can achieve a lower error rate with a smaller parameter size than ReCNAS can achieve because of the advantage of evolutionary algorithms in handling complex constraints. Note that in Table \ref {tab:2} and Table \ref{tab:3}, the optimal architecture searched by MetaNAS has fewer than 3M parameters, which means that the proposed method is suitable for scenarios with hardware resource constraints.

\section{Conclusion}\label{section:5} 
In this paper, we propose an innovative meta-knowledge-assisted EC-based NAS method, named MetaNAS, which can efficiently search robust architectures with high performance and favorable generalizability. We introduce a Meta-LR scheme into the search process. By pretraining with the training loss knowledge, a suitable LR schedule is obtained to guide the training process, which can provide flexibility to learn a suitable LR schedule when evaluating each individual. The Meta-LR scheme accommodates the complex search process because it brings prior knowledge into the training process, which expedites training convergence. An adaptive surrogate model is then used to evaluate architectures encoded by individuals. By predicting promising architectures using an adaptive threshold in few training epochs and then evaluating the selected architectures with complete training epochs, the number of sample architectures for the complete training epochs can be greatly reduced, and the search efficiency and generalizability can be improved on the premise of ensuring the effectiveness of the evaluation. Moreover, we design a period mutation operator to further increase population diversity and improve generalizability and robustness. The experimental results prove that the architecture of the proposed method is competitive with that of the state-of-the-art NAS methods.



\vspace{-1em}
\begin{IEEEbiography}
[{\includegraphics[width=1in,height=1.25in,clip,keepaspectratio]{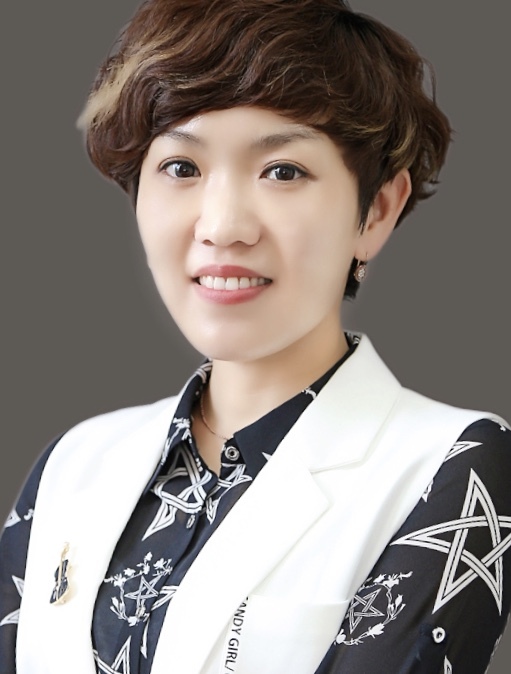}}]{Yangyang Li}
(Senior Member, IEEE) received B.S. and M.S. degrees in computer science and technology in 2001 and 2004, respectively, and a Ph.D. degree in pattern recognition and intelligent systems in 2007, all from Xidian University, Xi'an, China. She is currently a professor with the School of Artificial Intelligence, Xidian University. Her research interests include quantum intelligent computing and deep learning.
\end{IEEEbiography}

\begin{IEEEbiography}
[{\includegraphics[width=1in,height=1.25in,clip,keepaspectratio]{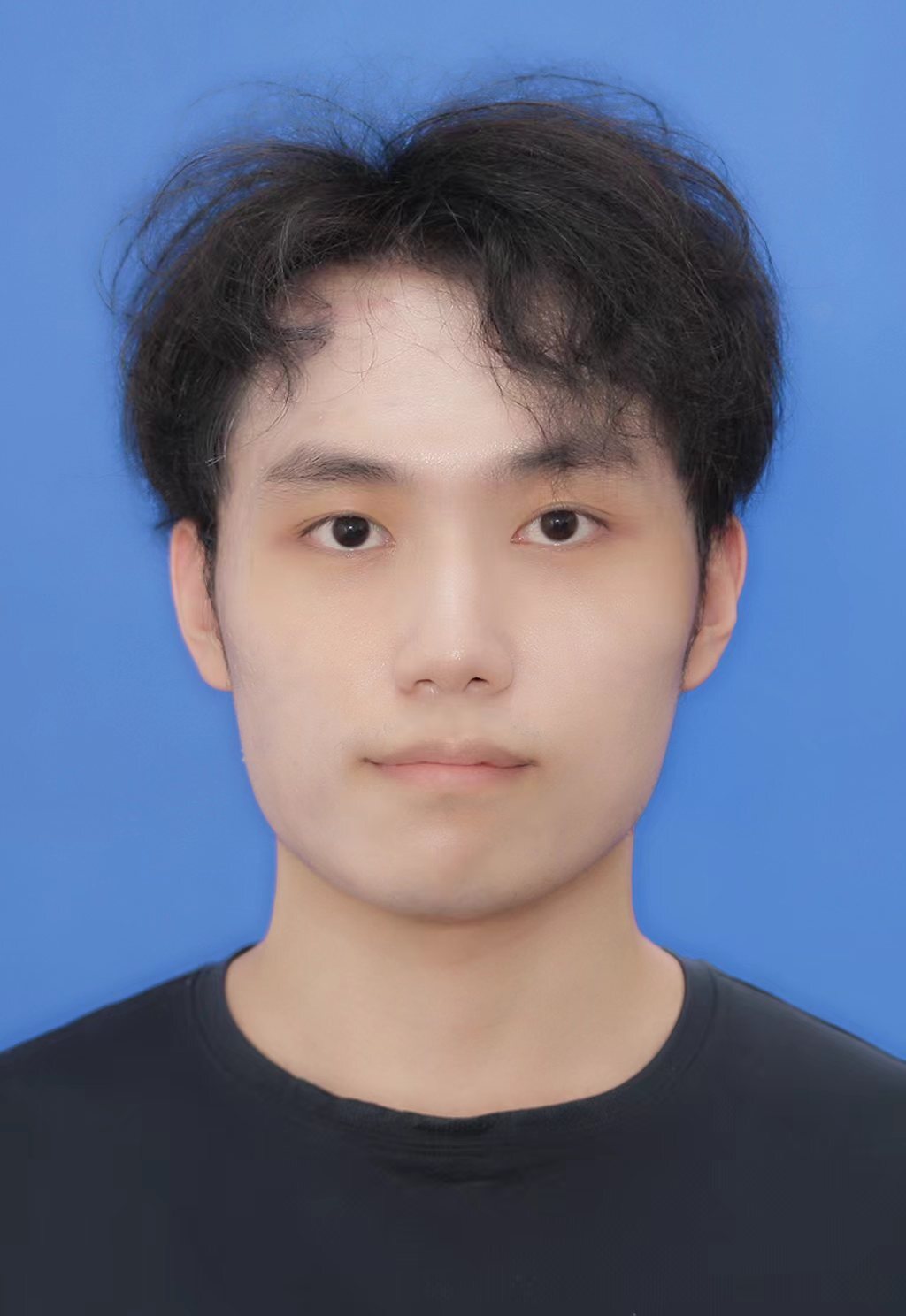}}]{Guanlong Liu}
received a B.E. degree from Jilin University, Changchun, China, in 2021. He is currently pursuing a Ph.D. degree in computer science at Xidian University, Xi'an, China. His current research interests include deep learning, evolutionary computation, and neural architecture search.
\end{IEEEbiography}

\begin{IEEEbiography}
[{\includegraphics[width=1in,height=1.25in,clip,keepaspectratio]{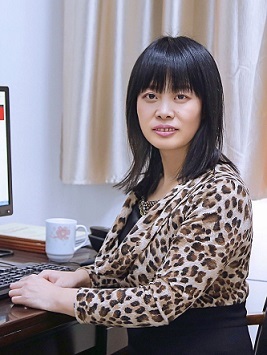}}]{Ronghua Shang}
(Senior Member, IEEE) received a B.S. degree in information and computation science and a Ph.D. degree in pattern recognition and intelligent systems from Xidian University, Xi'an, China, in 2003 and 2008, respectively. She is currently a professor at Xidian University. Her research interests include machine learning, pattern recognition evolutionary computation, image processing, and data mining.
\end{IEEEbiography}

\begin{IEEEbiography}
[{\includegraphics[width=1in,height=1.25in,clip,keepaspectratio]{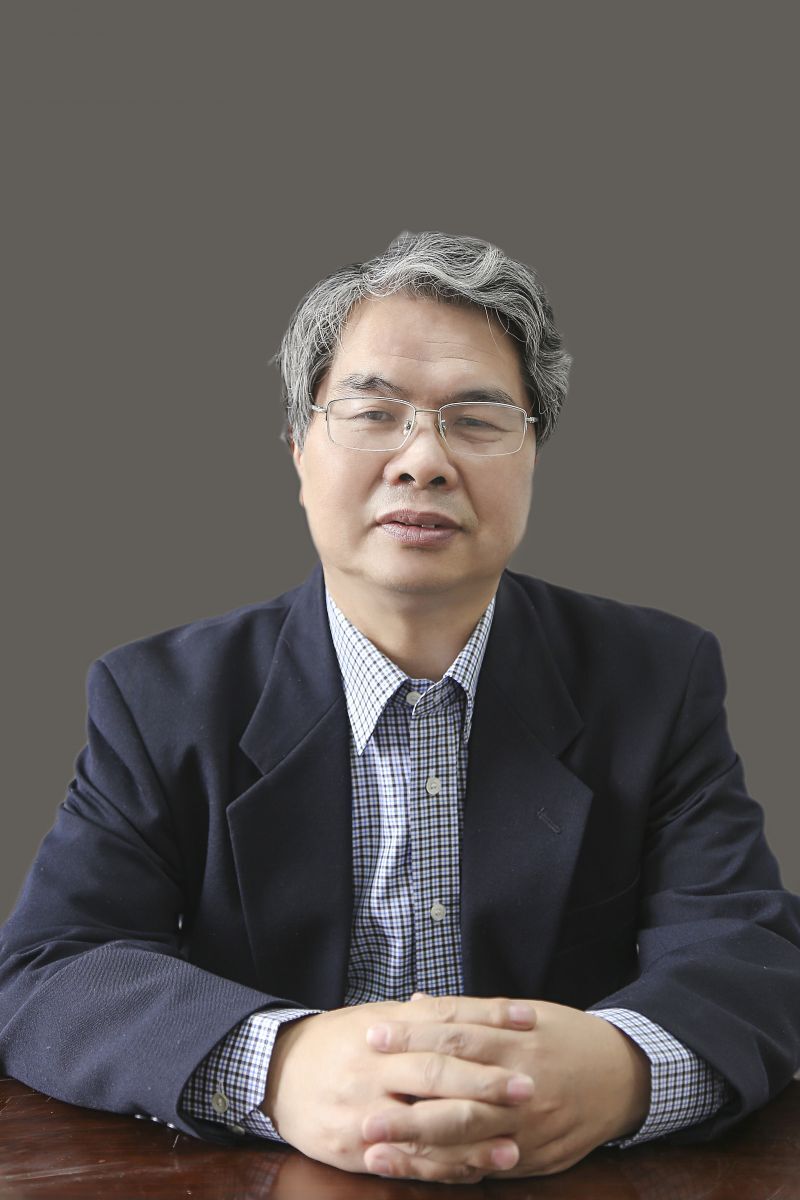}}]{Licheng Jiao}
(Fellow, IEEE) received a B.S. degree from Shanghai Jiaotong University, Shanghai, China, in 1982, and M.S. and Ph.D. degrees from Xi'an Jiaotong University, Xi'an, China, in 1984 and 1990, respectively, all in electronic engineering. From 1990--1991, he was a postdoctoral fellow at the National Key Laboratory for Radar Signal Processing, Xidian University, Xi'an. Since 1992, he has been a professor at the School of Electronic Engineering, Xidian University. He is currently the director of the Key Laboratory of Intelligent Perception and Image Understanding of the Ministry of Education of China, Xidian University. He is in charge of approximately 40 important scientific research projects and has authored or coauthored more than 20 monographs and 100 papers in international journals and conferences. His research interests include image processing, natural computation, machine learning, and intelligent information processing. Dr. Jiao is a member of the IEEE Xi'an Section Execution Committee and the Chairman of Awards and Recognition Committee, the Vice Board Chairperson of the Chinese Association of Artificial Intelligence, the Councilor of the Chinese Institute of Electronics, the Committee Member of the Chinese Committee of Neural Networks, and an expert of the Academic Degrees Committee of the State Council.
\end{IEEEbiography}

\end{document}